\documentclass[sigconf]{acmart}

\usepackage{booktabs} 

\usepackage{multirow}
\usepackage{xcolor,colortbl}
\usepackage{graphicx}
\usepackage{subcaption}
\usepackage{balance}
\usepackage{array}
\usepackage{booktabs}
\usepackage{multirow}
\usepackage{graphicx}
\usepackage{xcolor}
\usepackage{hyperref}
\usepackage{rotating}
\usepackage{tikz}
\usepackage{graphics}
\usepackage{enumitem}
\usepackage{balance}
\usepackage{amsmath}
\usepackage{comment}

\usepackage{xcolor,colortbl}

\newcommand{\baseline}[1]{\cellcolor{black}\textcolor{white}{{{#1}}}}

\copyrightyear{2025}

\acmArticle{4}
\acmPrice{15.00}

\author{Hugo Resende}
\affiliation{%
  \institution{Institute of Science and Technology, Universidade Federal de São Paulo}
  \city{São José dos Campos/SP} 
  \country{Brazil}
}
\email{hresende@unifesp.br}

\author{Isabela Borlido}
\affiliation{%
  \institution{Computer Science Department, Pontif\'icia Universidade Cat\'olica de Minas Gerais}
  \city{Belo Horizonte/MG}
  \country{Brazil}
}
\email{isabela.borlido@sga.pucminas.br}

\author{Victor Sundermann}
\affiliation{%
  \institution{Computer Science Department, Pontif\'icia Universidade Cat\'olica de Minas Gerais}
  \city{Belo Horizonte}
  \country{Brazil}
}
\email{vgmsundermann@sga.pucminas.br}

\author{Eduardo B. Neto}
\affiliation{%
  \institution{Institute of Science and Technology, Universidade Federal de São Paulo}
  \city{São José dos Campos/SP} 
  \country{Brazil}
}
\email{ebneto@unifesp.br}

\author{Silvio Jamil F. Guimar\~aes}
\affiliation{%
  \institution{Computer Science Department, Pontif\'icia Universidade Cat\'olica de Minas Gerais}
  \city{Belo Horizonte/MG}
  \country{Brazil}
}
\email{sjamil@pucminas.br}

\author{Fabio A. Faria}
\authornote{Currently, Dr. Fabio A. Faria is an assistant professor at the Instituto Superior Tecnico and  Collaborator Researcher at the Universidade Federal de São Paulo (UNIFESP).}
\affiliation{%
\institution{Instituto Superior Tecnico, Universidade de Lisboa}
\streetaddress{Av. Rovisco Pais 1}
 \city{Lisboa} 
 \country{Portugal}}
\email{fabio.faria@tecnico.ulisboa.pt}
 
\author{Álvaro L. Fazenda}
\affiliation{%
  \institution{Institute of Science and Technology, Universidade Federal de São Paulo}
  \city{São José dos Campos/SP} 
  \country{Brazil}
}
\email{alvaro.fazenda@unifesp.br}

\begin{document}

\title{Exploring Superpixel Segmentation Methods in the Context of Citizen Science and Deforestation Detection\\ \textcolor{red}{{(Complete Version)}}}

\begin{abstract}
Tropical forests play an essential role in the planet's ecosystem, making the conservation of these biomes a worldwide priority. However, ongoing deforestation and degradation pose a significant threat to their existence, necessitating effective monitoring and the proposal of actions to mitigate the damage caused by these processes. In this regard, initiatives range from government and private sector monitoring programs to solutions based on citizen science campaigns, for example. Particularly in the context of citizen science campaigns, the segmentation of remote sensing images to identify deforested areas and subsequently submit them to analysis by non-specialized volunteers is necessary. Thus, segmentation using superpixel-based techniques proves to be a viable solution for this important task. Therefore, this paper presents an analysis of 22 superpixel-based segmentation methods applied to remote sensing images, aiming to identify which of them are more suitable for generating segments for citizen science campaigns. The results reveal that seven of the segmentation methods outperformed the baseline method (SLIC) currently employed in the ForestEyes citizen science project, indicating an opportunity for improvement in this important stage of campaign development.  
\end{abstract}
%
%


\maketitle

\section{Introduction} 

The importance of tropical forests for global ecological balance is invaluable. These biomes harbor an impressive diversity of plant and animal species, many of which have not yet been discovered or studied. Additionally, tropical forests play an essential role in regulating the water cycle, maintaining periodic rainfall, and stabilizing the climate, even influencing global weather patterns. However, the increasing degradation of these forests due to fires, illegal logging, mining, and deforestation is causing irreparable damage. Recent research reports that in the Brazilian Legal Amazon, known as the largest tropical forest in the world, there were $2.5 \times 10^6 \ km^2$ of degraded area as of 2023~\cite{lapola2023drivers, reis2023econometric}. It is important to emphasize that the impacts of this degradation extend beyond the forest boundaries, affecting biodiversity, contributing to severe climate change, and threatening the livelihoods of local communities that depend on these ecosystems.

Aiming to propose solutions to combat deforestation and degradation of tropical forests, several monitoring programs have been designed and implemented~\cite{pro2023global, diniz2015deter}. Within the context of the Brazilian Legal Amazon, one of these programs is the Program for Monitoring Deforestation in the Legal Amazon by Satellite (PRODES), developed by the National Institute for Space Research (INPE) of Brazil~\cite{inpe-prodes}. PRODES utilizes a satellite monitoring system to provide annual estimates of deforestation rates in the Legal Amazon, allowing for the identification and mapping of deforested areas. Its significance lies in providing accurate and reliable data on deforestation, which are essential for guiding public policies, environmental enforcement, and conservation actions. By providing updated information on deforestation, PRODES plays a crucial role in protecting and preserving the tropical forests of the Legal Amazon.

Furthermore, various strategies, such as machine learning, image processing, and citizen science, have been utilized to identify deforestation areas. The ForestEyes project, for example, involves citizen science campaigns that actively engage non-specialist volunteers in monitoring tropical forests. Presently, these volunteers assist in labeling image segments that may contain deforested areas within the Brazilian Legal Amazon~\cite{dallaqua2019foresteyes,dallaqua2021foresteyes}. 
This project is composed of several interconnected modules, one of which is the machine learning module. Within this module, classification algorithms are trained using samples of forest and deforested areas, labeled based on classifications made by volunteers in the citizen science module. It is crucial to employ a robust segmentation method that generates easily understandable segments for untrained volunteers, with homogeneous and consistent pixels and well-defined boundaries, ensuring the best visual quality possible to maximize the accuracy of the volunteers.

\begin{figure}[t!]
\centering
\includegraphics[width=0.48\textwidth]{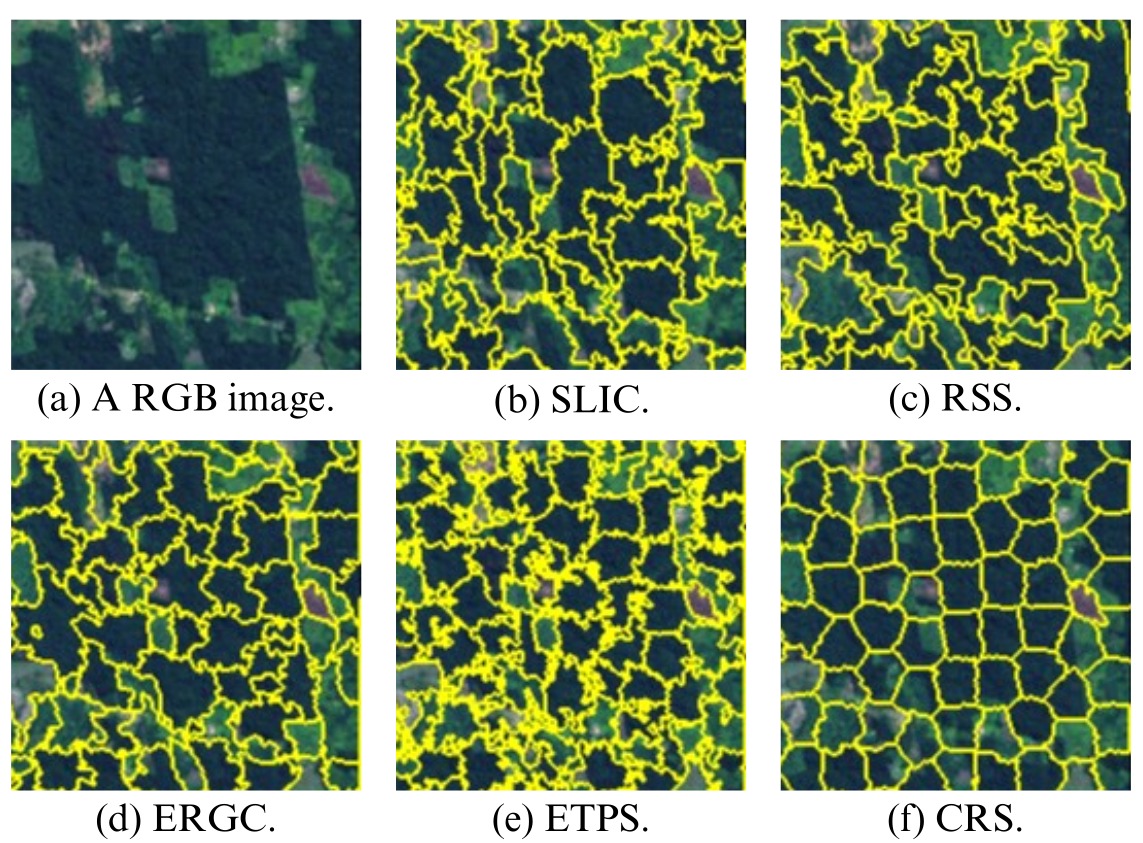}
\caption{Examples of segmentation results produced by SLIC, RSS, ERGC, ETPS and CRS methods.}
\label{fig:intro} 
\end{figure}

Presently, there is a considerable amount of methods employing different strategies to perform superpixel segmentation~\cite{barcelos2024review}.
Superpixels are regions of pixels that share characteristics such as color, texture, and intensity, and are used in several image processing applications. One of the benefits of superpixel segmentation compared to other approaches is the low computational complexity that most of its methods possess. 
They are also used to pre-process, reducing the data for further processing. 
For example, the Simple Linear Iterative Clustering (SLIC) method \cite{achanta2012slic} is known for its computational efficiency and ability to preserve object edges. 
However, it may encounter difficulties when segmenting images with complex textures or irregular shapes, resulting in inaccuracies in the segmentation results. 
Figure~\ref{fig:intro} presents a cropped remote sensing image, from study area 08 (SA-X08), composed of RGB bands, corresponding to a forested region (Figure \ref{fig:intro}(a)) and its respective segmentation result produced by SLIC (Figure \ref{fig:intro}(b)), RSS (Figure \ref{fig:intro}(c)), ERGC (Figure \ref{fig:intro}(d)), ETPS (Figure \ref{fig:intro}(e)) and CRS (Figure \ref{fig:intro}(f)) methods. This cropped, particularly, has a size of $170 \times 170$ pixels. In segmentation tasks, several irregular image regions corresponding to forest and deforestation regions are created, which can be used for citizen science campaigns in the ForestEyes project. 


In this sense, as SLIC is the segmentation algorithm adopted by the ForestEyes project, this study analyses $22$ leading superpixel-based segmentation algorithms available~\cite{barcelos2024review}. The evaluation focuses on the quality of the generated segments, aiming to determine their suitability for citizen science campaigns. Experimental results revealed that most of the analyzed methods outperformed SLIC's performance, suggesting possible contributions to enhance the project in question. 

In this work, our main contributions are threefold: (i) assessment of the performance of 22 different superpixel segmentation methods applied to remote sensing images to detect deforestation in tropical forests; (ii) a quantitative analysis of the segments created by superpixel methods; and (iii) proposal of a score for measuring the quality of the superpixel methods for the citizen science campaigns within the scope of the ForestEyes project, and identification of a new baseline.

The structure of this article is as follows: Section \ref{sec:background} discusses the fundamental concepts related to the quality of the segments and the segmentation methods studied. Section \ref{sec:foresteyes} presents the ForestEyes project. Section \ref{sec:experiments} addresses the presentation and discussion of the obtained results. Finally, Section \ref{sec:conclusion} presents the conclusions drawn from this research.

\section{Background}
\label{sec:background}
In this Section, the superpixel-based segmentation methods that are the subject of this study will be succinctly presented. Additionally, the superpixel evaluation measures will be explained, as well as those used to evaluate the methods from a citizen science perspective.

\subsection{Citizen Science}
Citizen Science (CS) can be defined as an approach that involves citizens in scientific activities such as data collection and analysis, allowing non-experts to collaborate alongside professional scientists \cite{haklay2021citizen}. 
 The advancements and increasing global popularity of remotely distributed ICT resources  (e.g., computers, tablets, and mobile phones) have helped the dissemination CS projects~\cite{SILVERTOWN2009}. Today, these projects are often referred to as Citizen Cyberscience and can be categorized into three main subtypes: Volunteered Computing, Volunteered Thinking, and Participatory Sensing~\cite{HAKLAY2013}.
In Volunteered Computing, a distributed processing power is provided by several volunteers donating their own computational resources.(SETI@home~\cite{SETI}, LHC@home~\cite{LHC2012} and climateprediction.net~\cite{CLIMATE2002}). Volunteered Thinking, involves participants utilizing their cognitive skills to analyze data or perform various tasks (Galaxy Zoo~\cite{AFW14}, volunteers~\cite{ZOONIVERSE2013},  and FoldIt~\cite{FOLDIT}). Meanwhile, in Participatory Sensing, the volunteers contribute with sensory information on a 
large-scale (eBird project~\cite{EBIRD2014}, CoralWatch~\cite{CORALWATCH}, and NoiseTube~\cite{NOISETUBE}).

CS projects create opportunities for volunteers to contribute to scientific advancements in areas ranging from environmental monitoring to disease research. One of the main objectives is to promote social inclusion in the production of scientific knowledge, expanding the reach of studies through public involvement ~\cite{silvertown2009new, lintott2011galaxy}. These initiatives have already proven to be effective in mobilizing volunteers for specific tasks such as Galaxy Zoo~\cite{lintott2011galaxy}, which uses public support to classify galaxies, and Science Scribbler: Placenta Profiles, focused on discovering patterns of anomalies in placental images, in order to explain how the placenta fails during pregnancy complications \cite{smith2023online}. Furthermore, in the environmental field, the ForestEyes project enlists citizens to identify deforested areas in the Amazon, combining human analysis with machine learning technologies to enhance detection accuracy. These campaigns demonstrate how large-scale collaboration can transform public engagement into valuable scientific data \cite{dallaqua2021foresteyes,fazenda_cacm2024}.

\subsection{Superpixel Methods}
\label{sec:superpixel_methods}

Superpixels are widely used in many applications to remove information redundancy, improve segmentation, and identify important image boundaries. However, most superpixel methods are designed for RGB images, thereby not handling hyperspectral information~\cite{subudhi2021survey}. Also, satellite images impose different challenges from natural ones, such as low-contrast regions and imbalanced data. Additionally, most works for hyperspectral image analysis with superpixels use classical superpixel methods instead of recent approaches~\cite{subudhi2021survey}. A survey for superpixel segmentation methods recently provided an extensive evaluation covering several methods, from recent and classic literature, proving their benefits and drawbacks~\cite{barcelos2024review}. 
Although the study in~\cite{barcelos2024review} provides useful insights, further analyses remain necessary to investigate which superpixel methods are more suitable to certain tasks and what and how much superpixel properties contribute to it. For instance, in Citizen Science projects, such as ForestEyes, untrained users play an important role in the process. Therefore, not only superpixels must produce good segmentation and classification of the deforested regions, but the segmented areas must be easy to interpret. In this sense, superpixels must have a good visual quality. In this work, we provide an in-depth assessment with 22 superpixel methods in remote sensing images for deforestation detection in tropical forests considering the ForestEyes project. 

Table~\ref{tab:spx_methods} presents the superpixel methods evaluated in this work and their categories. Additionally, we include a grid segmentation as a baseline. 
In the following, we briefly introduce superpixel methods used in this work and their categories~\cite{barcelos2024review}. 
Clustering-based methods, such as SLIC~\cite{achanta2012slic}, SCALP~\cite{GIRAUD-2018-SCALP}, and LSC~\cite{CHEN-2017-LSC}, perform an initial step to compute superpixel centers that subsequently conquer pixels based on a similarity function restricted to a spatial distance. These methods usually require post-processing to ensure connectivity. Dynamic-center-update clustering methods, such as SNIC~\cite{ACHANTA-2017-SNIC} and DRW~\cite{KANG-2020-DRW}, overcome this drawback by using a priority queue to process pixels and dynamically update superpixel centers. Conversely, boundary evolution-based methods, such as CRS~\cite{CONRAD-2013-CRS}, SEEDS~\cite{BERGH-2015-SEEDS}, ETPS~\cite{YAO-2015-ETPS}, and IBIS~\cite{BOBBIA-2021-IBIS}, adopt a coarse-to-fine strategy, iteratively updating superpixel labels of blocks of pixels. Similar to clustering-based ones, they may not guarantee connectivity or control over the number of superpixels. 

Boundary evolution-based methods are usually more efficient and produce compact superpixels, while dynamic-center-update clustering methods better manage the delineation compactness trade-off but are far less efficient. Path-based, hierarchical-based, and graph-based methods focus on object delineation and color homogeneity. These strategies model images as graphs. The former initializes pixel seeds that conquer the remaining pixels using path-based functions, in which the superpixels are the trees formed from this competition. ERGC~\cite{BUYSSENS-2014-ERGC}, ISF~\cite{VARGAS-2019-ISF}, RSS~\cite{CHAI-2020-RSS}, DISF~\cite{BELEM-2020-DISF}, ODISF~\cite{BELEM-2021-ODISF}, and SICLE~\cite{BELEM-2022-SICLE} are examples of path-based methods. On the other hand, hierarchical methodologies, such as in SH~\cite{WEI-2018-SH}, create a graph hierarchy from which one may compute many superpixel scales. 
Data distribution-based clustering methods assume that image pixels follow a certain distribution. For instance, the Gaussian distribution in GMMSP~\cite{BAN-2018-GMMSP}. Finally, deep-based methods employ clustering strategies in the network pipeline, performing hard or soft pixel-superpixel assignment. Although deep strategies are a trend in image processing, deep-based superpixel strategies, such as SSFCN~\cite{YANG-2020-SSFCN}, AINET~\cite{WANG-2021-AINET}, and SIN~\cite{YUAN-2021-SIN}, usually have poor performance. 


    
\begin{table}[t]
    \centering
    \caption{Superpixel methods and their categories.}     
    \label{tab:spx_methods}
    \begin{tabular}{ll}
        \toprule
         \textbf{Category} & \textbf{Methods} \\ \midrule
         Clustering & SLIC~\cite{achanta2012slic}, SCALP~\cite{GIRAUD-2018-SCALP}, LSC~\cite{CHEN-2017-LSC} \\
         Dynamic-center-update & SNIC~\cite{ACHANTA-2017-SNIC}, DRW~\cite{KANG-2020-DRW} \\
         Boundary evolution & \begin{tabular}[c]{@{}l@{}}CRS~\cite{CONRAD-2013-CRS}, SEEDS~\cite{BERGH-2015-SEEDS}, ETPS~\cite{YAO-2015-ETPS}, IBIS~\cite{BOBBIA-2021-IBIS} \\ \end{tabular} \\
         Path & \begin{tabular}[c]{@{}l@{}} ERGC~\cite{BUYSSENS-2014-ERGC}, ISF~\cite{VARGAS-2019-ISF}, RSS~\cite{CHAI-2020-RSS}, DISF~\cite{BELEM-2020-DISF}, \\ ODISF~\cite{BELEM-2021-ODISF}, SICLE~\cite{BELEM-2022-SICLE} \end{tabular} \\
         Hierarchical & SH~\cite{WEI-2018-SH} \\
         Graph & ERS~\cite{LIU-2011-ERS} \\
         Data distribution & GMMSP~\cite{BAN-2018-GMMSP} \\
         Deep & SSFCN~\cite{YANG-2020-SSFCN}, AINET~\cite{WANG-2021-AINET}, SIN~\cite{YUAN-2021-SIN} \\
         \bottomrule
    \end{tabular}
\end{table}


\subsection{Superpixel Evaluation Measures}
In superpixel segmentation, the adherence to the object's borders must ignore internal ground-truth region partitionings. \textbf{Boundary Recall (BR)}~\cite{martin2004learning} and \textbf{Undersegmentation Error (UE)}~\cite{neubert2012superpixel} are the most widely used metrics that evaluate boundary adherence. The former measures the recall of superpixel borders with the object borders. In BR, the boundary pixels are matched within a local neighborhood of size $(2r + 1)^2$, where $r$ is 0.0025 times the image diagonal. On the other hand, UE measures the superpixels' error according to ground-truth regions. In UE, a superpixel's error is based on the minimum overlap area with the ground-truth regions. 

\textbf{Similarity between Image and Reconstruction from Superpixels (SIRS)}~\cite{barcelos2022improving} evaluate color homogeneity in superpixels 
and provides more robustness to simple textures by modeling the color homogeneity problem as an image reconstruction problem. It first represents each superpixel as a small set of its most relevant colors. Then, it reconstructs the superpixel image by assigning each pixel the most similar color in its group. The reconstruction error is measured by the Mean Exponential Error (MEE) between the original and reconstructed image, and a Gaussian weighted at MEE defines the segmentation homogeneity. 

Finally, we evaluate superpixels' shape with \textbf{Compactness index (CO)}~\cite{schick2012measuring} and \textbf{Regularity (Reg)}. 
The former computes the similarity between the superpixel shape and a circle using the isoperimetric quotient, and the latter evaluates how much superpixels' areas differ in a segmentation, with the standard deviation of their areas. 

\subsection{Homogeneity Rate and Citizen Science Campaign-based Measures}

There are several metrics available to assess the quality of segmentations, among them, \textbf{Homogeneity Rate ($HoR$)} stands out, providing a measure of the degree of uniformity within each segment. This metric is calculated by determining the percentage of pixels belonging to the majority class in a binary segmentation. Originally conceived to differentiate whether a segment predominantly contained forest or deforestation pixels, $HoR$ is a valuable indicator of the internal coherence of each segment~\cite{dallaqua2021foresteyes}. 
The closer a segment's $HoR$ is to $1$, the better its quality, easily understood to be classified by volunteers. This indicates that the segment is highly homogeneous, with a clear predominance of pixels belonging to a single class, whether it be forest or deforestation. On the other hand, the closer a segment's $HoR$ is to $0.5$, the worse its quality, indicating a greater mixture of pixels from different classes. Segments with an $HoR$ close to $0.5$ are more difficult to classify by volunteers, as they exhibit questionable homogeneity and greater ambiguity regarding their true nature. 

\begin{figure}[ht!]
\centering
\includegraphics[width=0.45\textwidth]{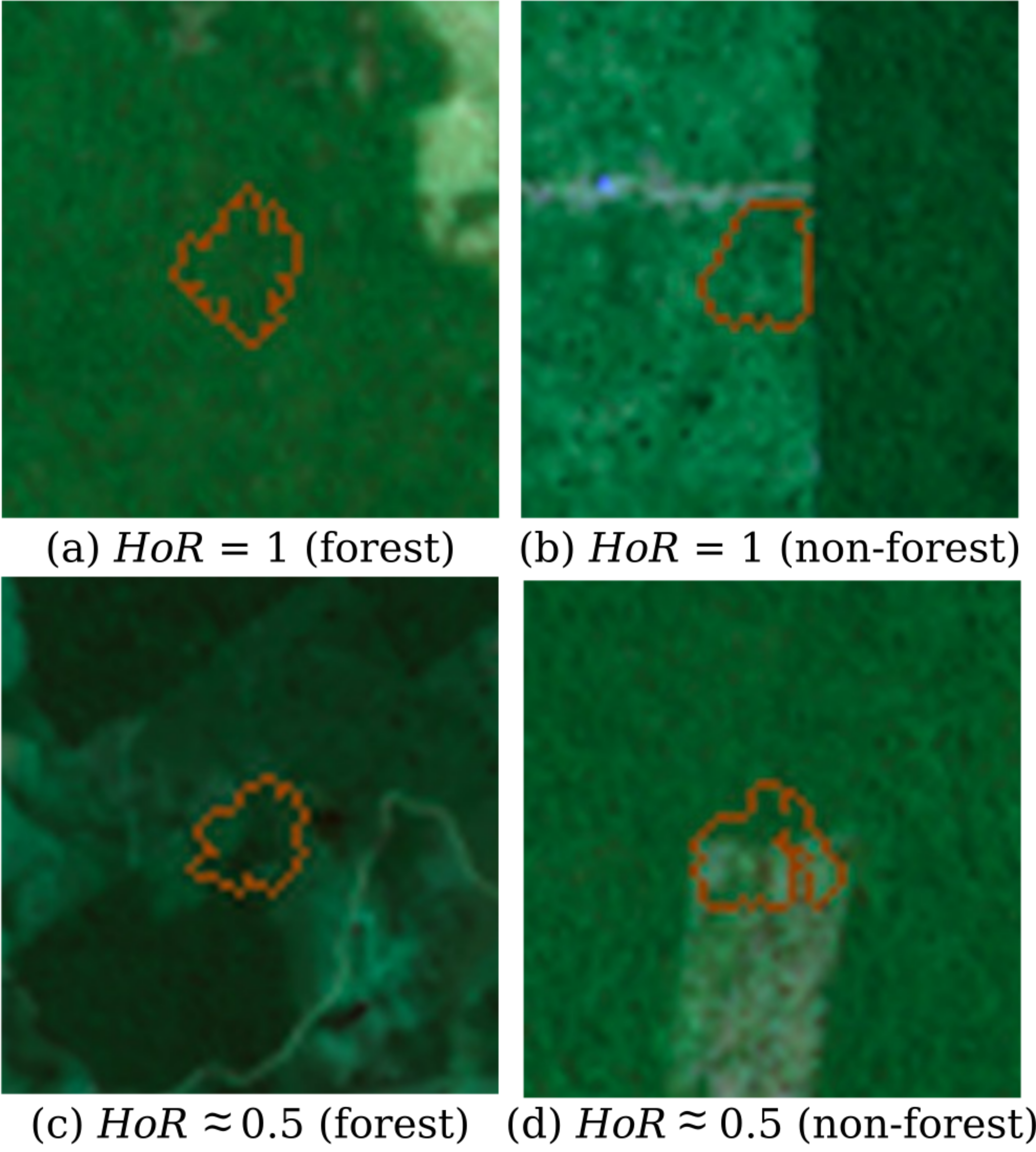}
\caption{Examples of segments with different $HoR$ values.}
\label{fig:hor}
\end{figure}

It is important to note that, while HoR is a valuable measure, highly irregular segments can also be problematic for classification by volunteers, as they may confuse volunteers and hinder the precise identification of the classes of interest. To illustrate these concepts, four examples of segments are presented in Figure \ref{fig:hor}. In the upper part of this image, there are two examples of segments with an $HoR$ equal to 1, one for the forest class (Figure \ref{fig:hor} (a)) and one for the non-forest class (Figure \ref{fig:hor} (b)). In the lower part, two examples with $HoR$ close to 0.5 can be observed (Figures \ref{fig:hor} (c) and (d)), indicating a considerable mixture of forest and deforestation classes. In Figure \ref{fig:hor} (d), particularly, it is also noticeable that the segment is quite irregular, which can hinder the analysis of volunteers if it is used in a citizen science campaign.

By computing $HoR$, it is possible to analyze and quantify the percentage of segments that can be used in citizen science campaigns, thereby evaluating the quality of each superpixel-based segmentation method tested in this study. Furthermore, $HoR$ allows for the derivation of various measures that enable analysis at both the segment and pixel levels. 

In this work, we proposed four measures to evaluate performance methods: 
(i) the percentage of so-called \textbf{Useful Segments (US)}, which have $HoR$ $\geq 0.7$ and a size of $70$ pixels or more \cite{dallaqua_grsl2020}; (ii) the percentage of \textbf{deforestation segments among the useful segments (DS)}; 
(iii) the percentage of segments with $HoR$ equal to $1$, referred to as \textbf{perfect $HoR$ (P\textit{HoR})}; and finally, 
(iv) the percentage of \textbf{pixels of the minority class (forest or non-forest) present in the useful segments}, representing the pixel-level error \textbf{(EP)}.

\section{ForestEyes: A Citizen Science Project for Deforestation Detection}
\label{sec:foresteyes}

\begin{figure*}
\centering
\includegraphics[width=1\textwidth]{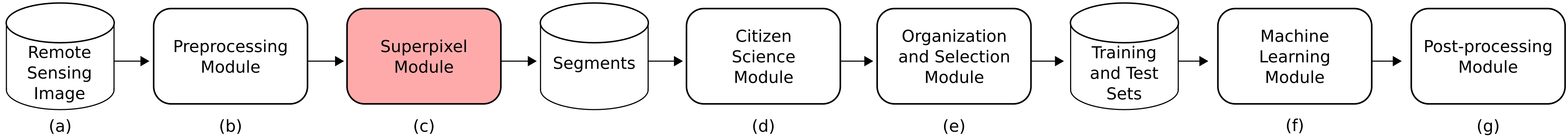}
\caption{The pipeline of the ForestEyes project. Highlighted is the target module of this work.}
\label{fig:pipeline} 
\end{figure*}

In this section, the ForestEyes project will be introduced, which consists of the modules shown in the pipeline of Figure \ref{fig:pipeline}. Each module will be further detailed in the following subsections.

\subsection{Acquisition of remote sensing images}

Initially, it is necessary to collect remote sensing data from the regions of interest that will be investigated. Currently, in the ForestEyes project, optical satellites such as Landsat-8 and Sentinel-2 are used. Landsat-8 remote sensing images are freely available via EarthExplorer\footnote{\url{https://earthexplorer.usgs.gov/}}. Although the satellite has 11 spectral bands, only 7 were utilized in this research (coastal aerosol (b1), blue (b2), green (b3), red (b4), near-infrared (nir - b5), shortwave infrared 1 (swir 1 - b6), and shortwave infrared 2 (swir 2 - b7)). On the other hand, Sentinel-2 is part of the European Space Agency’s (ESA) Copernicus program, comprising Sentinel-2A (launched in 2015) and Sentinel-2B (in 2017) \footnote{\url{https://dataspace.copernicus.eu/explore-data/data-collections/sentinel-data/sentinel-2}}. Both satellites are equipped with a Multispectral Instrument (MSI) that captures images across 13 bands, from visible to shortwave infrared. The visible bands offer a spatial resolution of 10 meters, while infrared bands provide 20 meters. Sentinel-2’s revisit frequency of 5 days offers critical data for environmental monitoring, agriculture, and disaster management \cite{drusch2012sentinel, main2017sen2cor}.

\subsection{Pre-processing Module} 
Once the remote sensing data is acquired, it is necessary to perform task such as resampling and cropping of the areas of interest, aligning them with the boundaries present in the official data provided by the Brazilian Amazon Forest Monitoring Program (PRODES). GIS software, such as QGIS, is used to efficiently handle large volumes of geospatial data during this process.

Additionally, an important preprocessing step is the extraction of Principal Component Analysis (PCA). PCA is a statistical technique that transforms correlated variables into uncorrelated ones, known as principal components. This reduces the data's dimensionality, preserving most of the original variability, which simplifies subsequent analyses and improves the efficiency of the algorithms applied in the project.

\subsection{Superpixel module} 

Once the images are properly processed and aligned with the PRODES ground truth, the segmentation process can begin. The ForestEyes project currently employs superpixel-based algorithms, given the need for a large number of segments across different classes, such as forest and recent deforestation, to generate the campaigns. During this phase, segmentations are evaluated based on various algorithm parameters, including the approximate number of segments per image and compactness.

Additionally, the quantity of useful segments (US) is carefully assessed for campaign creation. A class-wise analysis is conducted, using the PRODES data as an auxiliary resource. This step is crucial as it ensures that there are enough segments, often in the tens, to structure the campaigns. The availability of these segments is vital for providing a solid foundation for volunteer engagement and subsequent data collection.

\subsection{Citizen Science Module}

In the citizen science module, tasks are performed inside and outside the citizen science platform. Before launching the campaign, preliminary steps are required, such as defining false color compositions on which the segments will be projected and grouping the segments into multiple compositions to form the tasks. A workflow is then created on the citizen science platform (currently Zooniverse). This workflow includes configurable settings such as task layout (e.g., grid and flip), the expected number of responses per task, and the development of explanatory tutorials for volunteers. Therefore, given this sequence of activities, the importance of the citizen science module is evident, as more appropriate configurations can significantly impact subsequent modules in the project pipeline. In particular, a well-designed and structured campaign tends to improve volunteers' accuracy and, consequently, provide more reliable data for other components, such as the machine learning module.

\subsection{Organization and Selection Module}

The Organization and Selection module is composed of three main stages. The first stage involves classifying tasks according to their difficulty levels, determined through entropy based on the expected number of responses per task. The second stage conducts a thorough analysis of the volunteers' responses to identify the majority answer for each task, and with PRODES data, it calculates the volunteers' accuracy for each campaign. Lastly, the third stage selects samples for the Machine Learning module, using criteria that prioritize forest and deforestation segments, ensuring effective training for the model.

\subsection{Machine Learning Module}
The primary objective of the Machine Learning Module is to identify instances of deforestation within the selected segments. To accomplish this task, a machine learning technique might be adopted, such as Support Vector Machines (SVM)~\cite{svm}, k-Nearest Neighbors (kNN)~\cite{knn}, and Decision Trees (DT)~\cite{dt-survey}. In order to reduce the input data dimensionality and to extract the most relevant features from the segments, this study employed Haralick Texture Descriptors~\cite{HARALICK}. This descriptor is composed of $14$ features calculated from the gray-level co-occurrence matrix (GLCM). For this module, 13 of these features were computed using the Mahotas library~\cite{coelho2012mahotas} for each of the 3 image channels across 4 directions for all selected segments, following the experimental protocol for deforestation detection proposed by Dallaqua \textit{et al}~\cite{dallaqua_grsl2020}. These feature vectors serve as input for a machine-learning technique, enabling the detection and monitoring of deforestation patterns.


\subsection{Post-processing Module} This module is related to the analysis of achieved results. In the future, they might be used for making decisions such as a deforestation alert system and an auto-feedback scheme to automatically increase the training set.

\section{Experiments}
\label{sec:experiments}
This section presents the experimental methodology, experimental results, and their discussion.

\subsection{Experimental Methodology}
\label{sec:methodology}

The experimental database in this study comprised nine images from areas near the Xingu River Basin in the Brazilian Legal Amazon. These images were captured by the Landsat-8 satellite, operated by the United States Geological Survey (USGS), between July and October 2022. Table~\ref{tab:study_areas} presents the key characteristics of these study areas, including their extent in hectares and pixel dimensions. 
\begin{table}[ht!]
    \centering
    \caption{Main characteristics of the study areas collected by the Landsat-8 satellite.}
    \label{tab:study_areas}
    \resizebox{8.5cm}{!}{ 
    \begin{tabular}{|c|c|c|c|c|}
        \hline
        \textbf{Study Area} & \textbf{Extent (ha)} & \textbf{Acquisition Date} & \textbf{Width} & \textbf{Height} \\
        \hline \hline
        1 & 93,039.97 & 2022-09-03 & 1230 & 843 \\ \hline
        2 & 112,459.48 & 2022-09-03 & 1343 & 933 \\ \hline
        3 & 87,536.29 & 2022-08-19 & 977 & 998 \\ \hline
        4 & 60,733.32 & 2022-07-27 & 768 & 879 \\ \hline
        5 & 69,476.96 & 2022-07-27 & 928 & 833 \\ \hline
        6 & 152,865.48 & 2022-07-10 & 1788 & 950 \\ \hline
        7 & 77,884.28 & 2022-07-10 & 853 & 1017 \\ \hline
        8 & 92,591.06 & 2022-08-17 & 971 & 1064 \\ \hline
        9 & 104,845.23 & 2022-10-05 & 1047 & 1115 \\ 
        \hline
    \end{tabular}
    }
\end{table}

\begin{figure}[ht!]
\centering
\includegraphics[width=0.4\textwidth]{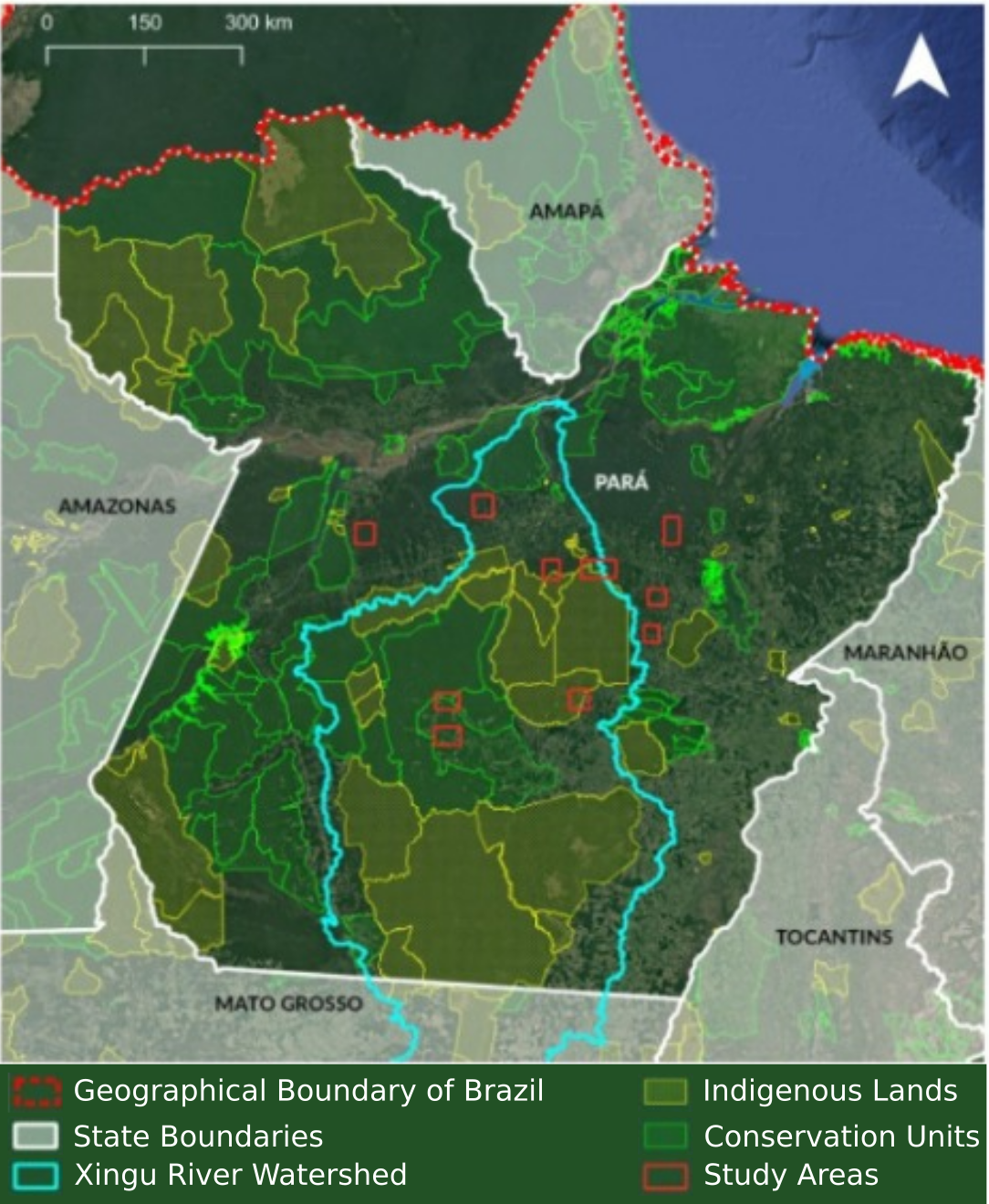}
\caption{Study areas located on Brazil's map.}
\label{fig:study_areas}
\end{figure}

Figure \ref{fig:study_areas} illustrates the study areas located in Pará, Brazil, highlighting key regions that require ongoing monitoring, such as indigenous lands and conservation units. The Xingu River Basin, rich in natural resources but vulnerable to illegal activities like mining, is given particular focus, which justifies its selection for this study.

Landsat-8 offers different spectral bands that can be combined for various tasks. In this research, we created images for segmentation using three principal components extracted from Principal Component Analysis (PCA). This decision was made based on Dallaqua et al.~\cite{dallaqua2019foresteyes}, which have performed experiments with the SLIC method, showing improved segmentation performance when using this band composition. 
The selection criterion for determining the band combination used 
was based on the percentage of useful segments (US). Dallaqua et al.~\cite{dallaqua_grsl2020} demonstrated that, from the perspective of human cognition, this type of segment is most suitable for citizen science campaigns in the ForestEyes project. Therefore, since a preliminary segmentation with the SLIC method showed that PCA was a viable option, we opted to use it to test the other segmentation methods. 

After constructing the images, all methods listed in Section~\ref{sec:superpixel_methods} were applied to the nine study areas. Most superpixel methods have parameters to set. As in~\cite{barcelos2024review}, to maintain fairness, we use the parameters recommended by the original authors. 
Due to computational resource constraints for executing the AINET method, we set the desired number of superpixels to $6000$ for all methods, which is the maximum feasible value for AINET in particular. Subsequently, when experimenting with each method, both superpixel evaluation metrics and those related to citizen science were computed for each study area. 

It is important to note that the values of $HoR$ were calculated based on ground-truth provided by the PRODES project, which provides a classification for each pixel in each study area. Additionally, in order to avoid generating very small segments and possibly minimize the percentage of useful segments, segments smaller than $70$ pixels were merged with their nearest neighbor. These mergers, while they may reduce the average $HoR$ value, tend to increase the quantity of useful segments.

\begin{table*}[ht!]
\centering
\caption{Rank for superpixel evaluation measures and Citizen Science strategy. On the right, is the final score of each method, in which lower values mean the best performance.}
\label{tab:res:rank}
\begin{tabular}{@{}llccccccc|lccccc|lc@{}}
\toprule
\multirow{2}{*}{\textbf{Method}} &  & \multicolumn{7}{c|}{\textbf{Superpixel evaluation}} &  & \multicolumn{5}{c|}{\textbf{Citizen Science Strategy}} &  & \multirow{2}{*}{\begin{tabular}[c]{@{}c@{}}\textbf{Final}\\ \textbf{Score}\end{tabular}} \\
 &  & DS & BR & UE & SIRS & CO & Reg & \textbf{Score$_{SP}$} &  & DS & US & P$HoR$ & EP & \textbf{Score$_{CS}$} &  &  \\ \cmidrule(r){1-1} \cmidrule(lr){3-9} \cmidrule(lr){11-15} \cmidrule(l){17-17} 
RSS &  & 2 & 11 & 9 & 7 & 16 & 7 & 8.667 &  & 7 & 17 & 2 & 9 & 8.800 &  & 8.733 \\
ERGC &  & 5 & 8 & 15 & 10 & 11 & 11 & 10.000 &  & 12 & 12 & 2 & 6 & 8.000 &  & 9.000 \\
ETPS &  & 5 & 2 & 19 & 2 & 12 & 20 & 10.000 &  & 22 & 9 & 4 & 1 & 9.000 &  & 9.500 \\
CRS &  & 15 & 6 & 22 & 5 & 5 & 21 & 12.333 &  & 16 & 2 & 1 & 8 & 6.800 &  & 9.567 \\
LSC &  & 12 & 4 & 14 & 6 & 18 & 15 & 11.500 &  & 8 & 16 & 7 & 2 & 8.300 &  & 9.900 \\
SH &  & 3 & 12 & 12 & 4 & 17 & 6 & 9.000 &  & 4 & 18 & 4 & 17 & 10.800 &  & 9.900 \\
GMMSP &  & 10 & 1 & 18 & 1 & 19 & 18 & 11.167 &  & 3 & 22 & 7 & 7 & 9.800 &  & 10.483 \\
\baseline{SLIC} &\baseline{ } & \baseline{5} & \baseline{10} & \baseline{7} & \baseline{12} & \baseline{14} & \baseline{13} & \baseline{10.167} &\baseline{ }  & \baseline{17} & \baseline{11} & \baseline{12} & \baseline{5} & \baseline{11.300} &\baseline{ }  & \baseline{10.733} \\
DISF &  & 1 & 5 & 8 & 3 & 20 & 4 & 6.833 &  & 1 & 20 & 20 & 18 & 14.800 &  & 10.817 \\
ISF &  & 3 & 13 & 10 & 8 & 15 & 5 & 9.000 &  & 6 & 15 & 17 & 13 & 12.800 &  & 10.900 \\
SNIC &  & 9 & 7 & 13 & 11 & 9 & 16 & 10.833 &  & 18 & 10 & 7 & 10 & 11.300 &  & 11.067 \\
ERS &  & 16 & 3 & 21 & 9 & 8 & 19 & 12.667 &  & 9 & 14 & 4 & 11 & 9.500 &  & 11.083 \\
SSFCN &  & 18 & 15 & 20 & 13 & 4 & 10 & 13.333 &  & 14 & 8 & 11 & 4 & 9.300 &  & 11.317 \\
SIN &  & 10 & 18 & 11 & 17 & 2 & 8 & 11.000 &  & 11 & 4 & 15 & 20 & 12.500 &  & 11.750 \\
IBIS &  & 8 & 14 & 6 & 15 & 7 & 14 & 10.667 &  & 20 & 7 & 13 & 12 & 13.000 &  & 11.833 \\
AINET &  & 16 & 17 & 16 & 14 & 6 & 9 & 13.000 &  & 21 & 6 & 13 & 3 & 10.800 &  & 11.900 \\
SCALP &  & 19 & 16 & 17 & 16 & 3 & 12 & 13.833 &  & 19 & 3 & 7 & 15 & 11.000 &  & 12.417 \\
SEEDS &  & 13 & 9 & 4 & 18 & 10 & 17 & 11.833 &  & 15 & 5 & 21 & 14 & 13.800 &  & 12.817 \\
GRID &  & 20 & 19 & 5 & 19 & 1 & 22 & 14.333 &  & 13 & 1 & 18 & 16 & 12.000 &  & 13.167 \\
DRW &  & 13 & 20 & 1 & 21 & 13 & 3 & 11.833 &  & 10 & 13 & 22 & 19 & 16.000 &  & 13.917 \\
ODISF &  & 22 & 22 & 2 & 22 & 21 & 1 & 15.000 &  & 2 & 19 & 16 & 22 & 14.800 &  & 14.900 \\
SICLE &  & 21 & 21 & 3 & 20 & 22 & 2 & 14.833 &  & 5 & 21 & 18 & 21 & 16.300 &  & 15.567 \\ \bottomrule
\end{tabular}
\end{table*}

\subsection{Results and Discussion} 
\label{sec:results}


In this section, we evaluated the performance of 22 superpixel methods in remote sensing images, focused on deforestation. Our main goal is to identify the best superpixel-based segmentation method to support the development of citizen science campaigns within the ForestEyes project. Therefore, to achieve this, we created two analyses and rankings to identify the best superpixel methods. 
The first assesses the method's performance based on classical superpixel analysis (see on the left side of the Table \ref{tab:res:rank}, named Superpixel evaluation), while the second evaluates the methods using citizen science-related criteria (see on the right side of Table 
\ref{tab:res:rank}, named Citizen Science Strategy). For each evaluation measure, we calculated the arithmetic mean of the values obtained for the nine study areas (RoIs) for both superpixels and citizen science. 

We compute ranking scores for superpixels (Score$_{SP}$) and citizen science (Score$_{CS}$), as follows. First, we compute the arithmetic mean of the results for the nine study areas (RoIs). Then, we rank methods for each metric, in which smaller ranks indicate better performance. The (Score$_{SP}$) and (Score$_{CS}$) values are the arithmetic mean of the individual rankings for each algorithm, where lower scores indicate superior performance (score columns in Table~\ref{tab:res:rank}). 
Finally, we compute the Final Score (last column in Table \ref{tab:res:rank}) as the arithmetic mean of both scores (Score$_{SP}$ and Score$_{CS}$), in which a lower score indicates superior performance.  In both evaluations, DS (Deforestation Segments among the useful segments) is the score for non-forest regions only. Also, DS is used as a tiebreaker criterion. 

\subsubsection{\textbf{Superpixel Evaluation:}} 

As one may see in Table~\ref{tab:res:rank}, DISF, ISF, RSS, and SH provide the best delineation (BR and UE). GMMSP, ETPS, DISF, SH, and CRS create more homogeneous superpixels, while GRID, SIN, SCALP, SSFCN, and CRS exhibit better compactness. Regarding non-forest regions only, DISF, RSS, SH, and ISF perform better. It's important to note that these methods prioritize delineation, which may result in poorer compactness. Considering all measures for superpixel evaluation, DISF, RSS, SH, ISF, and ERGC have superior performance, i.e., lower Score$_{SP}$ values. Among these methods, ERGC may have more leakage (UE), DISF could produce less compact superpixels, and ISF might have inferior delineation. 

\subsubsection{\textbf{Citizen Science Strategy:}} 

As observed in Table \ref{tab:res:rank} for Citizen science ranking, CRS achieved the highest ranking for this evaluation. Although it did not perform well in generating a high percentage of deforestation segments (DS) among the useful segments, it excelled in generating useful segments overall (forest and non-forest) and perfect segments (P$HoR$). Conversely, in superpixel evaluation, CRS has the worst UE, producing superpixels with more leakage and less regularity. Due to this performance, it was the fourth according to the Final Score. Conversely, RSS cannot produce compact superpixels, while ERGC produces less homogeneous superpixels. Also, both ERGC and ETPS are less focused on compactness than CRS.

\subsubsection{\textbf{The Best Superpixel Methods for Citizen Science:}} 

After constructing the performance rankings of the methods, based on superpixel evaluation measures and citizen science metrics, a final ranking was computed (Final Score -- last column in Table~\ref{tab:res:rank}). So this ranking was calculated based on the arithmetic mean of the methods' rankings in the two previous rankings (Score$_{SP}$ and Score$_{CS}$). Furthermore, this Final Score allows observing the presence of seven segmentation methods (RSS, ERGC, ETPS, CRS, LSC, SH, and GMMSP) showing a better quantitative performance than the baseline method currently used in the ForestEyes Project (SLIC method -- black color row).

\section{Conclusions}
\label{sec:conclusion}
In this study, we assessed the performance of 22 different superpixel-based segmentation methods using nine remote sensing images to detect deforestation in tropical forests. We also conducted a quantitative analysis of the segments created by each method, aiming to find a better alternative to the baseline method (SLIC). This resulted in a ranked list based on measures closely related to citizen science campaigns within the scope of the ForestEyes project. Combining the superpixel methodology with citizen science strategies, we produced a final list of the best methods among all 22 tested in this paper. As future work, we plan to use the top five methods (RSS, ERGC, ETPS, CRS, and LSC) to create Citizen Science campaigns in the ForestEyes project, comparing their effectiveness with the baseline method (SLIC). Additionally, the 5 classification databases will serve as training sets for machine learning classification across the larger area covered by other segments in the test set, which will be also evaluated. We also plan to analyze deeply how superpixel methods perform in deforestation classification only, especially for recently deforested areas. Identifying which satellite bands perform better in superpixel segmentation, especially in recent deforestation regions, is another issue to be addressed. Recently, several deep segmentation networks have been proposed, we also plan to evaluate them in the same context.

\bibliographystyle{ACM-Reference-Format}
\bibliography{bibs} 

\end{document}